\documentclass[10pt, letterpaper, conference]{IEEEtran}
\IEEEoverridecommandlockouts
\usepackage{cite}
\usepackage{amsmath,amssymb,amsfonts}
\usepackage{algorithmic}
\usepackage{graphicx}
\usepackage{textcomp}
\usepackage{xcolor}
\usepackage{stfloats}
\usepackage{gensymb}
\usepackage{subfigure}
\usepackage{booktabs}
\usepackage[ruled,vlined]{algorithm2e}
\usepackage{makecell}
\usepackage{enumerate}
\def\BibTeX{{\rm B\kern-.05em{\sc i\kern-.025em b}\kern-.08em
    T\kern-.1667em\lower.7ex\hbox{E}\kern-.125emX}}
\begin{document}

\title{Real-time End-to-End Federated Learning:\\ An Automotive Case Study}


\author{
    \IEEEauthorblockN{Hongyi Zhang\IEEEauthorrefmark{1}, Jan Bosch\IEEEauthorrefmark{1}, Helena Holmström Olsson\IEEEauthorrefmark{2}}
    \IEEEauthorblockA{\IEEEauthorrefmark{1}\textit{Chalmers University of Technology}, Gothenburg, Sweden. \\Email: \{hongyiz, jan.bosch\}@chalmers.se} 
    
    \IEEEauthorblockA{\IEEEauthorrefmark{2}\textit{Malmö University}, Malmö, Sweden. \\Email: helena.holmstrom.olsson@mau.se}
}

\maketitle

\begin{abstract}

With the development and the increasing interests in  ML/DL fields, companies are eager to apply Machine Learning/Deep Learning approaches to increase service quality and customer experience. Federated Learning was implemented as an effective model training method for distributing and accelerating time-consuming model training while protecting user data privacy. However, common Federated Learning approaches, on the other hand, use a synchronous protocol to conduct model aggregation, which is inflexible and unable to adapt to rapidly changing environments and heterogeneous hardware settings in real-world scenarios. In this paper, we present an approach to real-time end-to-end Federated Learning combined with a novel asynchronous model aggregation protocol. Our method is validated in an industrial use case in the automotive domain, focusing on steering wheel angle prediction for autonomous driving. Our findings show that asynchronous Federated Learning can significantly improve the prediction performance of local edge models while maintaining the same level of accuracy as centralized machine learning. Furthermore, by using a sliding training window, the approach can minimize communication overhead, accelerate model training speed and consume real-time streaming data, proving high efficiency when deploying ML/DL components to heterogeneous real-world embedded systems.

\end{abstract}
\vspace{5pt}

\section{Introduction}

\begin{figure}[htbp]
    \centering
    \includegraphics[scale=0.26]{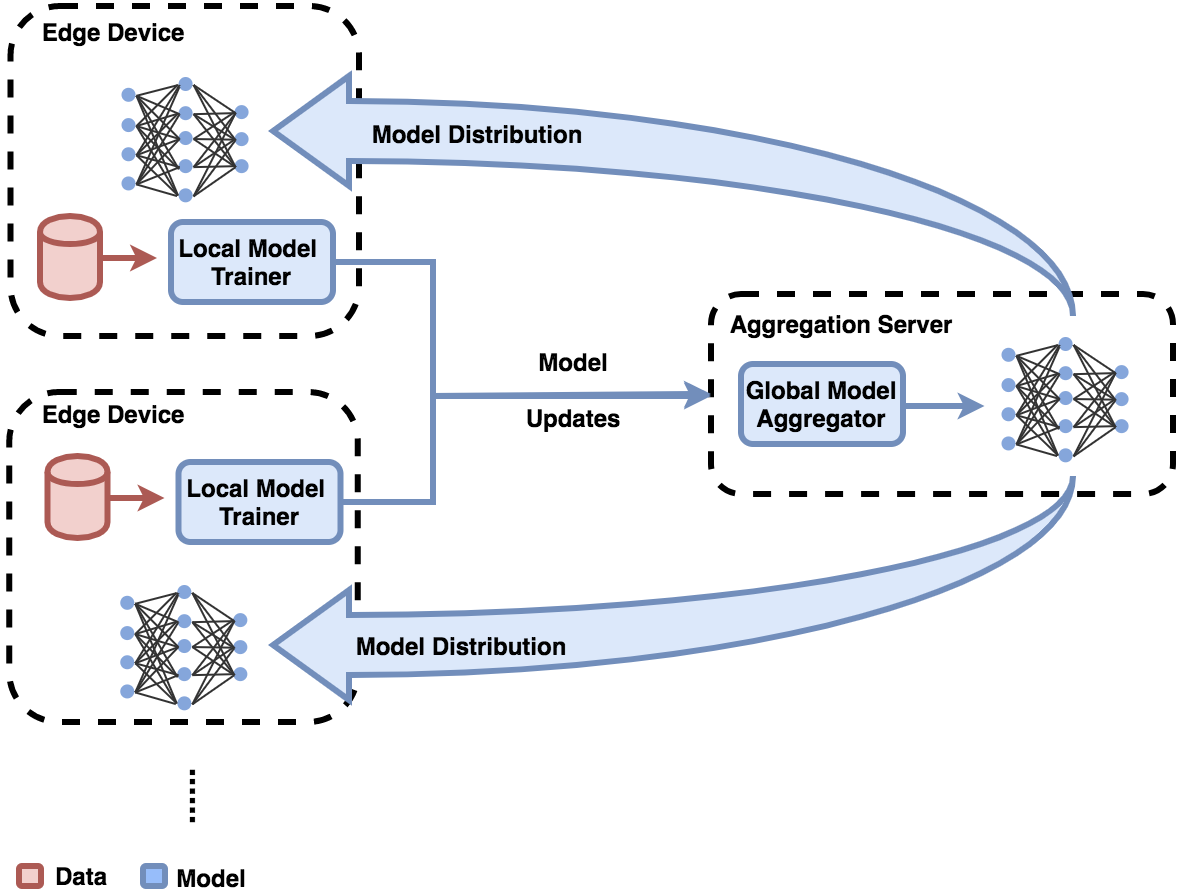}
    \caption{A typical Federated Learning System is depicted in the diagram. The light blue components are related to the model, while the red components are related to the data.}
    \label{fig:fl}
\end{figure}

With the development of distributed edge computer computing and storage capabilities, using computation resources on the edge becomes a viable option \cite{compute}. Federated Learning has been adopted as a cost-effective solution due to its model-only sharing and parallel training characteristics. A simple diagram of a Federated Learning system is shown in Figure \ref{fig:fl}. Local model training is carried out in this framework, and data generated by edge devices do not need to be shared. Weight updates are instead sent to a central aggregation server, which generates the global model. The method overcomes the shortcomings of the conventional centralized Machine Learning approach, which only conducts model training on a single central server, such as data privacy, massive bandwidth costs, and long model training time.

This paper builds on our previous research, ``End-to-End Federated Learning for Autonomous Driving Vehicles'' \cite{end-to-end}, in which we discovered that Federated Learning can significantly reduce model training time and bandwidth consumption. However, with the synchronous aggregation protocols used in our previous research and current Federated Learning applications and analysis, such as FedAvg \cite{fedavg}, we realized that it is difficult for businesses to incorporate Federated Learning components into their software products \cite{li2020federated}. Until model aggregation, a synchronous aggregation protocol requires the server to wait for all of the edge devices to complete their training rounds. Since real-world systems may include heterogeneous hardware configurations and network environments \cite{park2016learning}, the aggregation server cannot expect all participating edge devices to upload their local models at the same time. The situation will become worse and unmanageable with the increasing number of edge devices. Furthermore, our previous research also identified the challenges of deploying AI/ML components into a real-world industrial context. As J. Bosch et al. defined in \textit{"Engineering AI Systems: A Research Agenda"} \cite{bosch2020engineering}, AI engineering refers to AI/ML-driven software development and deployment in production contexts. We found that the transition from prototype to the production-quality deployment of ML models proves to be challenging for many companies \cite{mlchallenge} \cite{lwakatare2019taxonomy}.

Therefore, in order to put Federated Learning into effect, in this paper, we present a novel method for consuming real-time streaming data for Federated Learning and combining it with the asynchronous aggregation protocol. This paper makes three contributions. First, we employ Federated Learning, a distributed machine learning technique, and validate it with a key automotive use case, steering wheel angle prediction in the field of autonomous driving, which is also a classic end-to-end learning problem. Second, we present a real-time end-to-end Federated Learning method for training Machine Learning models in a distributed context. Third, we empirically evaluate our approach on the real-world autonomous driving data sets. Based on our findings, we show the effectiveness of our method over other methods of learning, including the common synchronous Federated Learning approach.

The remainder of this paper is structured as follows. 
 Section \ref{sec: method} details our research method, including the simulation testbed, the utilized machine learning method and the evaluation metrics. Section \ref{sec:stru} presents the real-time end-to-end Federated Learning approach utilized in this paper. Sections \ref{sec:eval} evaluates our proposed learning method to empirical data sets.  Finally, Section \ref{sec:concl} presents conclusions and future work.

\section{Related Work}
\label{sec:related}

\subsection{Steering Wheel Angle Prediction}

One of the first pioneer research of utilizing the neural network for steering wheel angle prediction is described in \cite{pomerleau1998autonomous}. The author used pixel information from simulated road images as inputs to predict steering command, which proves that a neural network is able to perform steering angle prediction from image pixel values. Recently, more advanced networks are utilized to predict the steering angles. H. M. Eraqi et al. propose a convolutional long short-term memory (c-LSTM) to learn both visual and dynamic temporal dependencies of driving, which demonstrate more stable steering by 87\% \cite{eraqi2017end}. Shuyang et al. \cite{du2019self} designed a 3D-CNN model with LSTM layers to predict steering wheel angles.

The concept of end-to-end learning was first proposed in \cite{bojarski2016end}, where authors built and constructed a deep convolutional neural network to directly predict steering wheel angles and monitor the steering wheel. In this research, ground truth was directly captured from real-time human behaviour. Their methods demonstrate that a convolutional neural network can learn steering wheel angle directly from input video images without the need for additional road information such as road marking detection, semantic analysis, and so on. In order to enhance model prediction accuracy, we use a two-stream model in our approach. Due to its robustness and lower training cost as compared to other networks such as DNN \cite{haq2020comparing}, 3D-CNN \cite{du2019self}, RNN \cite{eraqi2017end}, and LSTM \cite{valiente2019controlling}, the model was first proposed in \cite{simonyan2014two} and applied in \cite{fernandez2018two}. However, previous research for this use case has concentrated primarily on the training model in a single-vehicle. We will use Federated Learning in this paper to accelerate model training speed and boost model quality by forming a global awareness of all edge vehicles.

\subsection{Federated Learning in Automotive}

The automotive industry is a promising platform for implementing Machine Learning in a federated manner. Machine learning models can be used to forecast traffic conditions, identify pedestrian behaviour, and assist drivers in making decisions \cite{vecomputer}\cite{peddetect}. However, since vehicles must have an up-to-date model for safety purposes, Federated Learning has the potential to accelerate Machine Learning model development and deployment while protecting user privacy \cite{konevcny2016federated1}.

On top of Federated Learning, Lu et al. \cite{lu2019collaborative} test the failure battery for an electric vehicle. Their methods demonstrate the efficacy of privacy serving, latency reduction, and security protection. Saputra et al. \cite{saputra2019energy} forecast the energy demand for electric vehicle networks. They dramatically minimize the bandwidth consumption and efficiently protect sensitive user information for electric vehicle users by using Federated Learning. Samarakoon et al. \cite{samarakoon2018federated} propose a distributed approach to joint transmit power and resource allocation in vehicular networks that enable low-latency communication. When compared to a centralized approach, the proposed method can reduce waiting queue length without increasing power consumption and achieve comparable model prediction efficiency. Doomra et al. \cite{doomra2020turn} present a Federated Learning-trained long short-term memory (LSTM)-based turn signal prediction (on or off) model. All of these approaches, however, are faced with synchronous aggregation protocols that are unsuitable for real-world heterogeneous hardware. As a result, in this paper, we present an asynchronous aggregation protocol combined with Federated Learning and validate it with one of the most essential use cases in the automotive industry.

\section{Method}
\label{sec: method}

The analytical technique and research method mentioned in \cite{zhang2003machine} were used in this study to conduct a quantitative measurement and comparison of real-time Federated Learning and conventional centralized learning methods. The article presents some recommendations for applying machine learning methods to software engineering activities, as well as methods for demonstrating how they can be conceived as learning problems and addressed in terms of learning algorithms. The mathematical notations, testbed and hardware configuration, convolutional neural network, and evaluation metrics used to solve the problem of steering wheel angle prediction are presented in the following sections.

\subsection{Mathematical Notations}

The mathematical notations that will be used in the paper are introduced here first:

\bgroup
\def\arraystretch{1.5}
\begin{tabular}{p{1.1in}p{1.9in}}

$\displaystyle A_t$ & An image frame matrix at time $t$\\
$\displaystyle O_t=f({A_t, A_{t-1}})$ & An optical-flow matrix at time $t$\\
$\displaystyle \theta_t$ & Steering wheel angle at time $t$\\
$\displaystyle \hat{\theta_t}$ & Predicted steering wheel angle at time $t$\\

\end{tabular}
\egroup

\subsection{Data Traces and Testbed}

The datasets used in this paper are from the SullyChen collection of labelled car driving data sets, which can be found on Github under the tag \cite{sullychen}. To conduct our experiments, we chose Dataset 2018 from this collection. The dataset contains various driving data such as road video clips, steering angle on roads, and so on. Dataset 2018 is 3.1 GB in size and contains approximately 63,000 files. This dataset tracks a 6-kilometer path along the Palos Verdes Peninsula in Los Angeles. Our experiment datasets were chosen from the first 40,000 image frames.

\begin{figure*}[!t]%
\centering
\subfigure[][Vehicle 1: Highway \& City]{%
\label{fig:ex3-a}%
\includegraphics[height=1.8in]{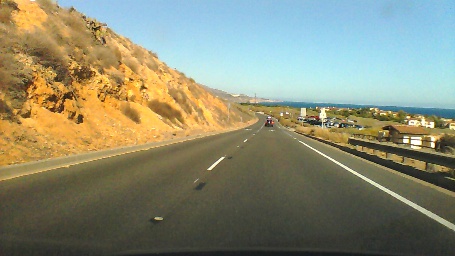}}%
\hspace{25pt}%
\subfigure[][Vehicle 2: Highway \& City]{%
\label{fig:ex3-b}%
\includegraphics[height=1.8in]{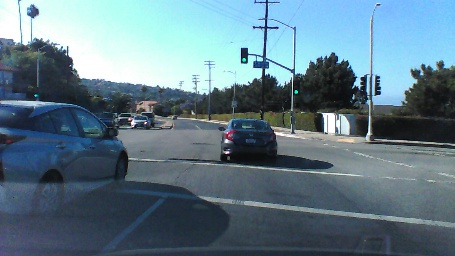}} \\
\subfigure[][Vehicle 3: Hill]{%
\label{fig:ex3-c}%
\includegraphics[height=1.8in]{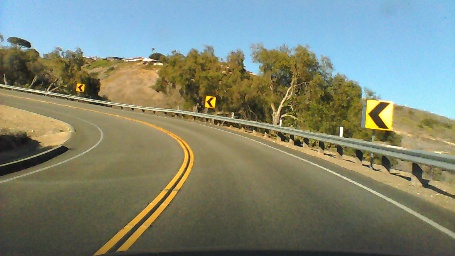}}%
\hspace{25pt}%
\subfigure[][Vehicle 4: Hill \& City]{%
\label{fig:ex3-d}%
\includegraphics[height=1.8in]{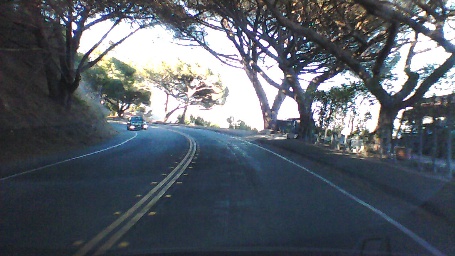}}%
\caption[]{Driving scenarios in each edge vehicle.}%
\label{fig: sce}%
\end{figure*}

The data streams were simulated on four edge vehicles to provide a comprehensive evaluation. The data was divided into four sections and distributed to edge vehicles prior to our simulation. In each edge vehicle, the first 70\% of data are considered input streaming driving information that was used for model training, while the remaining 30\% are potential stream information. As shown in Figure \ref{fig: sce}, training datasets for each edge vehicle in our experiment include a variety of driving scenarios.

\begin{figure}[h]%
\centering
\subfigure[][Vehicle 1]{%
\label{fig:ex3-a}%
\includegraphics[height=1in]{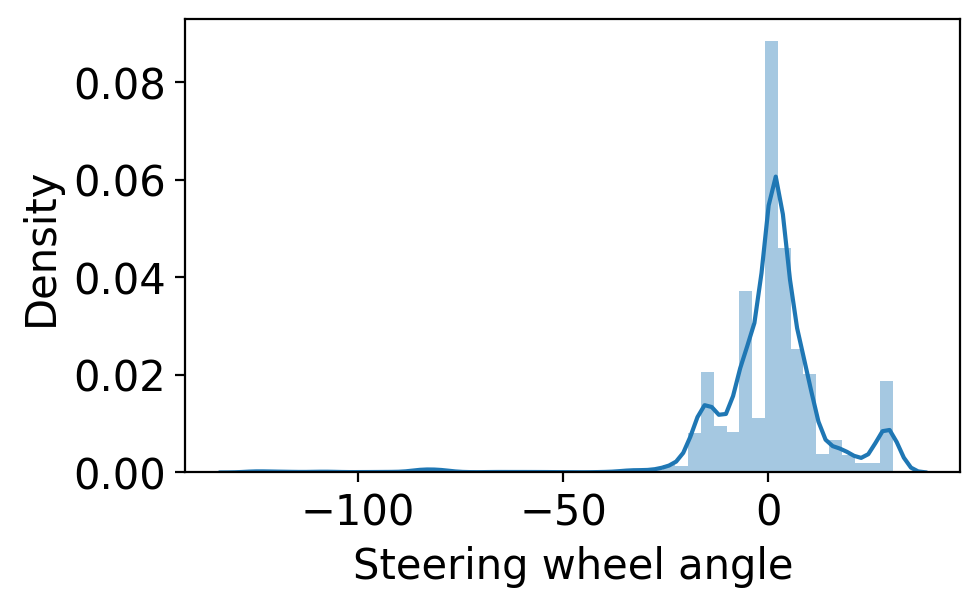}}%
\hspace{8pt}%
\subfigure[][Vehicle 2]{%
\label{fig:ex3-b}%
\includegraphics[height=1in]{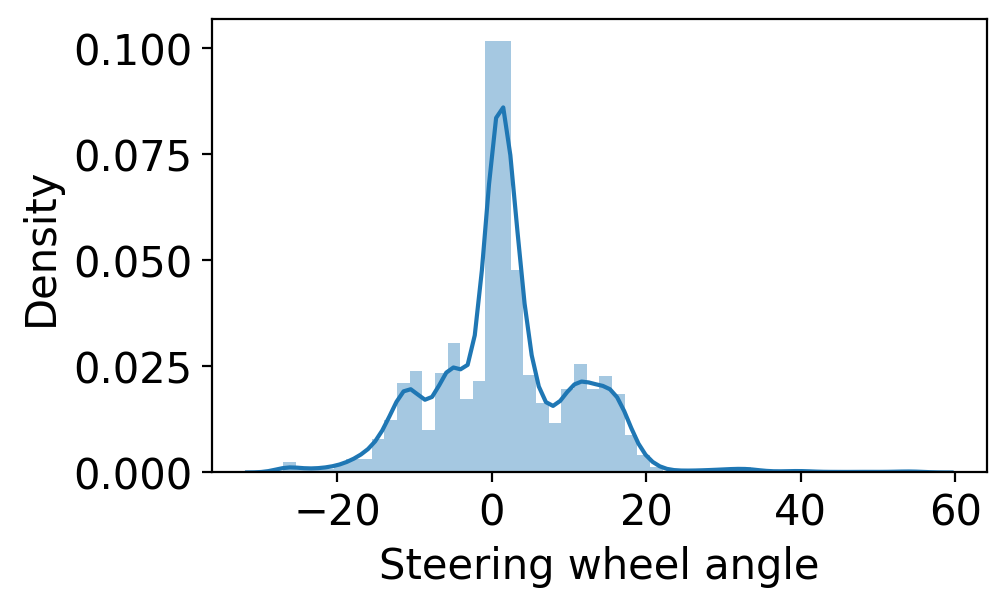}} \\
\subfigure[][Vehicle 3]{%
\label{fig:ex3-c}%
\includegraphics[height=1in]{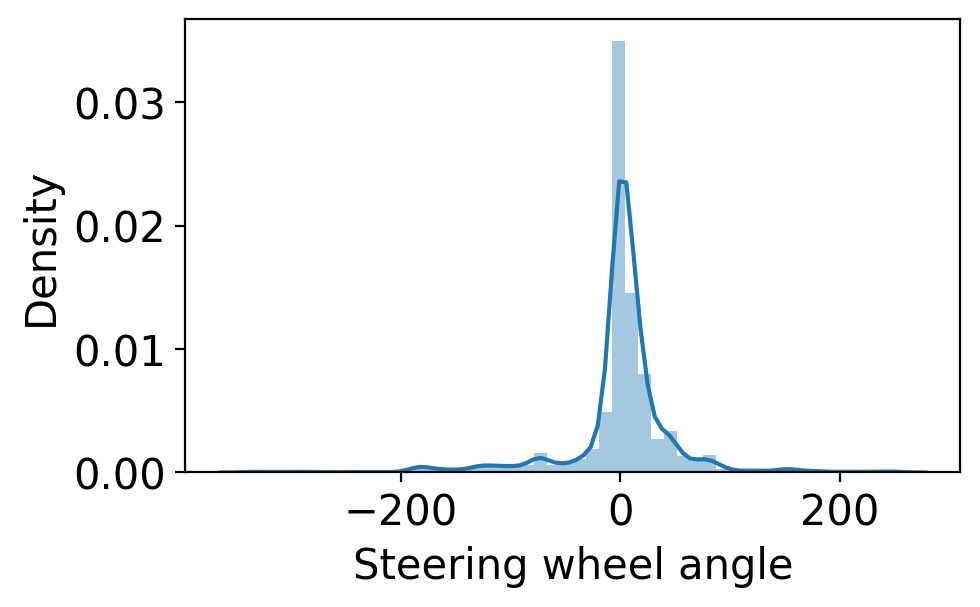}}%
\hspace{8pt}%
\subfigure[][Vehicle 4]{%
\label{fig:ex3-d}%
\includegraphics[height=1in]{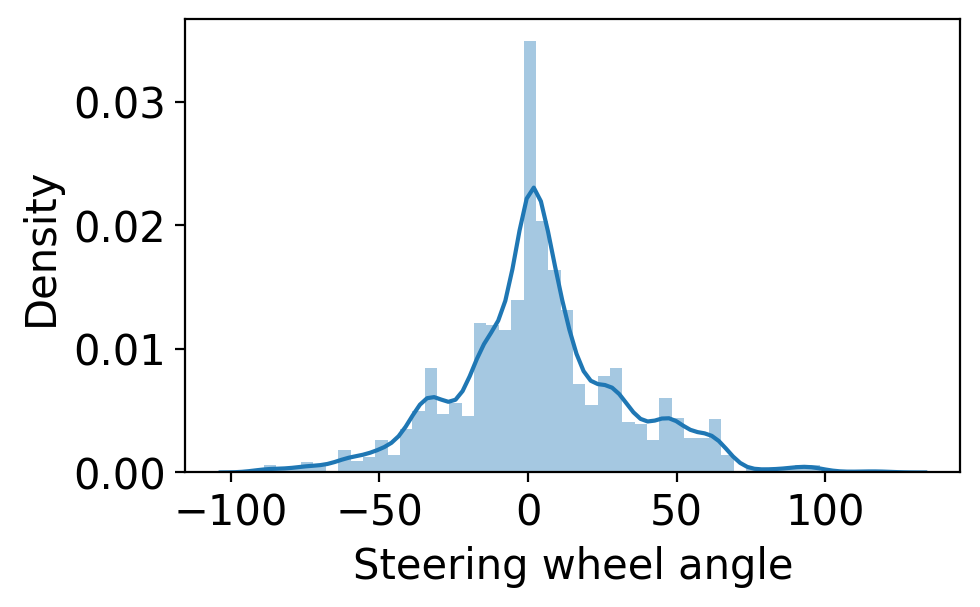}}%
\caption[]{Data distribution in each edge vehicle.}%
\label{fig:density-result}%
\end{figure}

The data distribution in each edge vehicle is depicted in Figure \ref{fig:density-result}. When driving on a hill, the steering wheel angles have a greater range than when driving on a highway or in a neighbourhood. The majority of driving angles in edge vehicles 1 and 2 falls within the range $[-50\degree, 50\degree]$, while in edge vehicles 3 and 4, the range is $[-100\degree, 100\degree]$. The graph shows that when driving on a hill, vehicles may encounter more turns than when driving on a highway or in a city.

The models were continuously trained based on the recorded data and used future streaming driving data to perform prediction and validation on the steering wheel angle information.

The hardware information for all of the servers is given in table \ref{tab:hard}. To simulate aggregation and edge functions, one of the five servers was designated as the aggregation server, while the others operated as edge vehicles. In order to simulate a heterogeneous edge area, GPU settings were only available in Vehicles 1, 3, and 4 (Vehicle 1: Tesla V100, Vehicle 3, 4: Tesla T4).

\begin{table}[htbp]
\caption{Hardware setup for testbed servers}
\label{tab:hard}
\centering

\begin{tabular}{c|c}
\hline
CPU       & Intel(R) Xeon(R) Gold 6226R          \\ \hline
Cores     & 8                                    \\ \hline
Frequency & 2.90 GHz                             \\ \hline
Memory    & 32 GB                                \\ \hline
OS        & Linux 4.15.0-106-generic             \\ \hline
GPU  & \makecell{Nvidia Tesla V100 GPU (Edge vehicle 1)\\Nvidia Tesla T4 GPU (Edge vehicle 3, 4)} \\ \hline
\end{tabular}

\end{table}

\subsection{Machine Learning Method}
\label{sec:ml}

\begin{figure*}[t]
  \begin{center}
    \includegraphics[scale=0.4]{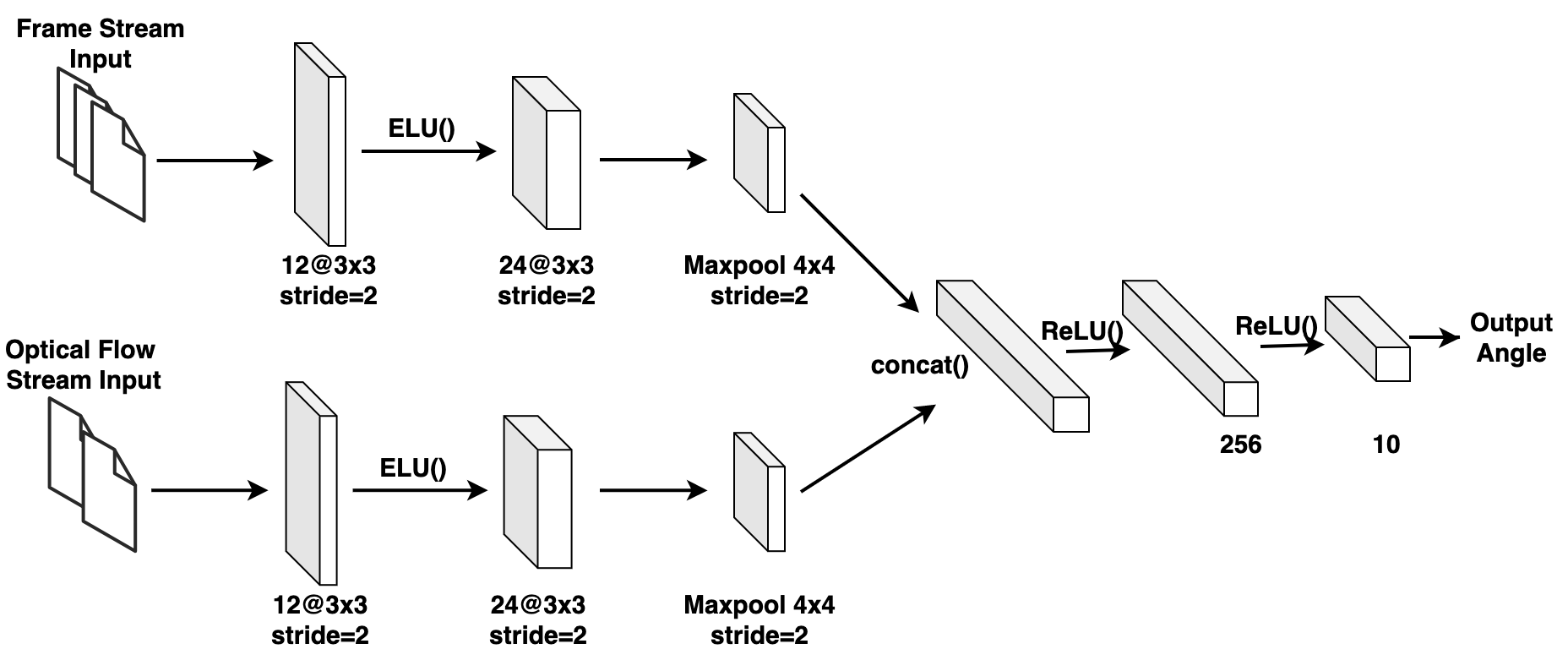}
    \caption{The two input branches each have two 3x3 convolution layers in a convolutional neural network. The first layer has 12 output channels that are enabled with the ELU function, while the second layer has 24, which is then followed by 4x4 max pooling. All with stride values of 2 or higher. With the ReLu activation, there are two completely linked layers with 250 and 10 units after concatenating two branches.}
    \label{fig:model}
  \end{center}
\end{figure*}

In this paper, steering wheel angle prediction is performed using a two-stream deep Convolutional Neural Network (CNN) \cite{simonyan2014two} \cite{fernandez2018two}. The architecture is described in detail in Figure \ref{fig:model}. Each stream in our implementation has two convolutional layers and a max-pooling layer. After concatenating, there are two fully-connected layers activated by the ReLU function.

The model has two distinct neural branches that take spatial and temporal information as inputs to two streams and then output the expected steering angle. The model consumes three frames of RGB images for the first stream, which can be denoted as $\{A_{t-2}, A_{t-1}, A_{t}\}$. The second stream is a two-frame optical flow measured from two consecutive frames $O_{t-1} = f(\{A_{t-2}, A_{t-1}\})$ and $O_t = f(\{A_{t-1}, A_{t}\})$.

\begin{figure*}[b]%
\centering
\subfigure[$A_{t-1}$]{%
\label{fig:first}%
\includegraphics[height=1.2in]{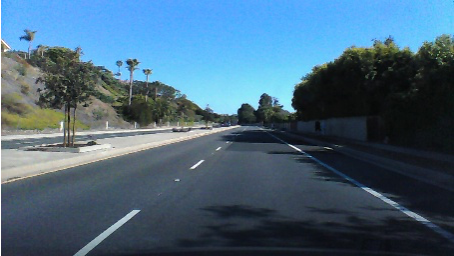}}%
\qquad
\subfigure[$A_t$]{%
\label{fig:second}%
\includegraphics[height=1.2in]{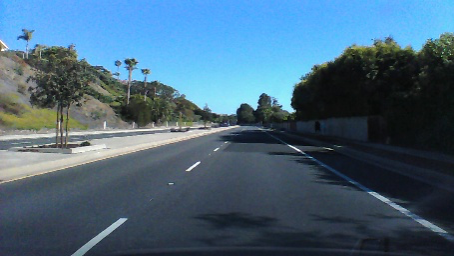}}%
\qquad
\subfigure[$O_t=f(\{A_{t-1}, A_{t}\})$]{%
\label{fig:second}%
\includegraphics[height=1.2in]{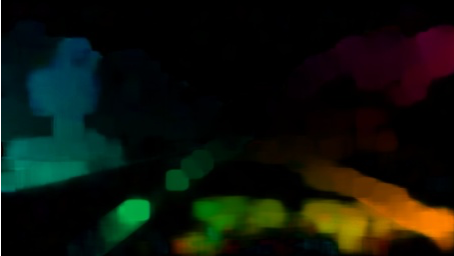}}%
\qquad
\caption{Example of the optical flow (a) Previous Frame (b) Current Frame (c) Optical flow of current vision frame.}
\label{fig:optical}
\end{figure*}

Optical flow is a typical temporal representation in video streams that captures the motion differences between two frames \cite{horn1981determining}. The optical flow calculation method used in this paper is based on Gunnar Farneback's algorithm, which is implemented in OpenCV \cite{farneback2003two}. Figure \ref{fig:optical} shows an example optical flow matrix created by two consecutive image frames.


The aim of training a local convolutional neural network is to find the model parameters that result in the smallest difference between the prediction and ground truth steering angles. As a result, we choose mean square error as the local model training loss function in this case:
\begin{equation} 
Loss=\frac{1}{N}\sum^N_{t=1}(\theta_t-\hat{\theta_t})^2
\end{equation}

Here, $N$ represents the batch size while $\theta_t$ and $\hat{\theta_t}$ represent the ground truth and the predicted steering wheel angle value at time $t$. During the process of model training in each edge vehicle, all the image frames will be firstly normalized to $[-1, 1]$. The batch size is 16 while the learning rate is set to $10^{-5}$. The optimizer utilized is Adam \cite{adam}, with parameters $\beta_1=0.6$, $\beta_2=0.99$ and $\epsilon=10^{-8}$.

\subsection{Evaluation Metrics and Baseline Model}
\label{sec:metric}

We chose three metrics and three baseline models in order to provide fruitful outcomes and assessment.  The three metrics include angle prediction performance, model training time and bandwidth cost:

\begin{itemize}

\item \textbf{Angle prediction performance}: Root mean square error (RMSE), a common metric for measuring the difference between prediction results and ground truth. The metrics will provide a reasonable estimate of the trained model's quality in each edge vehicle.

\item \textbf{Model training time}: The total time cost for training a model at the edge vehicles is known as this metric. As a consequence, the average of four edge vehicles is obtained. This metric shows the pace at which local edge devices update their model, which is critical for systems that need to evolve quickly in order to adapt to a rapidly changing environment. By testing the model deployment timestamp, the metrics were calculated in all of the vehicles.

\item \textbf{Bandwidth cost}: The total number of bytes transmitted during the entire training procedure is known as this metric. This metric shows the overall cost of communication resources needed to achieve an applicable convolutional neural model.

\end{itemize}

The three baseline models include models trained by applying the traditional centralized learning approach, the locally trained model without model sharing and the Federated Learning with the synchronous aggregation protocol:

\begin{itemize}

\item \textbf{Traditional Centralized Learning model (ML)}: This baseline model was trained using a centralized learning method, which is still widely used in current machine learning research and software applications. All data from edge vehicles is collected to a single server prior to model training. The hyper-parameters of this model training are identical to those of Federated Learning, as described in section \ref{sec:ml}. The results can then be compared to models trained using Federated Learning techniques.

\item \textbf{Locally trained model without model sharing (Local ML)}: 

Each edge vehicle is used to train this baseline model. In contrast to Federated Learning, no models will be exchanged during the training process. The prediction accuracy can be applied to the Federated Learning model to see if Federated Learning outperforms those independently trained local models.

\item \textbf{Synchronous Federated Learning (FL)}: 
FedAvg is the algorithm applied here. It is a synchronous method that is widely used in Federated Learning research. Before aggregating global models, the server has to wait for all participants to finish updating their local models.

\end{itemize}

\section{Real-time End-to-End Federated Learning}
\label{sec:stru}

\begin{figure*}[t]
  \begin{center}
    \includegraphics[scale=0.5]{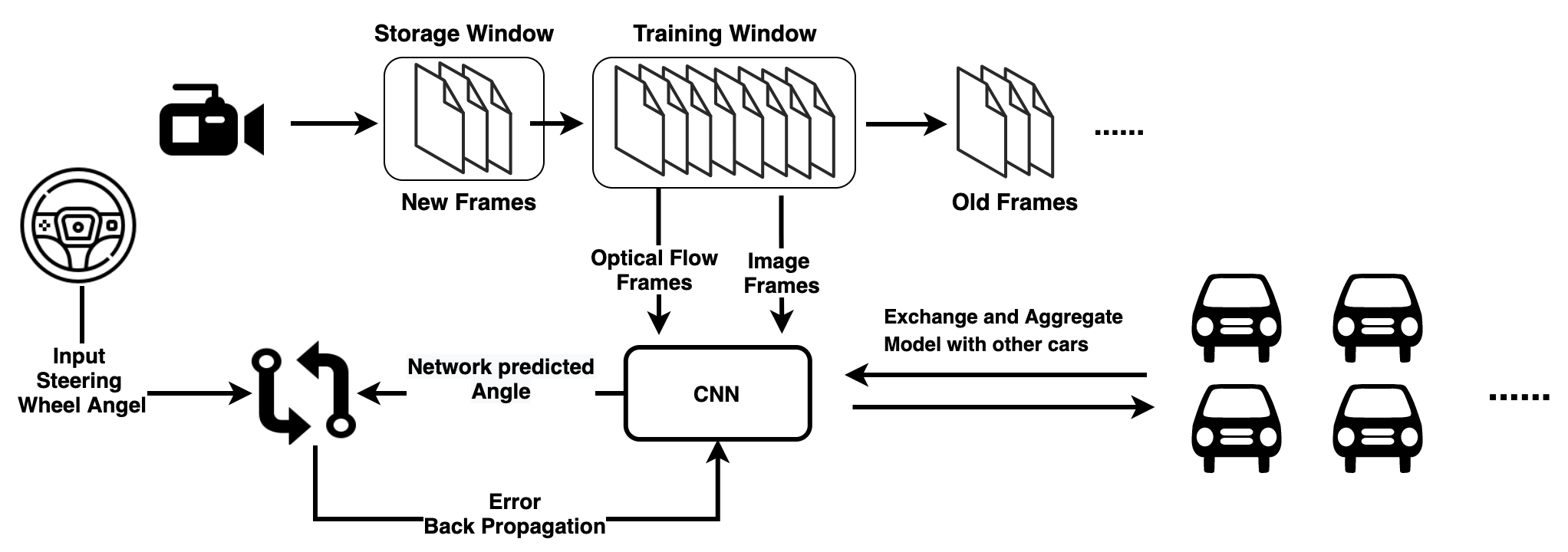}
    \caption{Diagram of real-time end-to-end Federated Learning in a single vehicle.}
    \label{fig:train_edge}
  \end{center}
\end{figure*}

\begin{algorithm}[]
\SetAlgoLined
\DontPrintSemicolon

    \SetKwFunction{FMain}{Server\_Function} 
    \SetKwProg{Fn}{Function}{:}{}  
    \Fn{\FMain{}}{
        initialize $w_0$
        
        initialize $ver \longleftarrow a_l$
        
        \While {True} {

            {$ w_{t+1}^{k}, ver_k \longleftarrow Client\_Update(w_t, ver);$ }
            \vspace{3pt}
            
            {$ w_{t+1} \longleftarrow (1 - \alpha) \times w_t + \alpha \times w_{t+1}^k$ }
            
            {where $\alpha = \frac{1}{ver-ver_k+1};$}
            \vspace{3pt}
            
            {$ver \longleftarrow ver + 1;$}
            } 
        
}
\textbf{End Function}

        \vspace{5pt}

    \SetKwFunction{FMain}{Client\_Update} 
    \SetKwProg{Fn}{Function}{:}{}  
    \Fn{\FMain{w, ver}}{
        $ \beta \longleftarrow $(split $P_k$ into batches of size $B$);
        
        \While{True} {
        
        \For{each local epoch $i$ from 1 to E } {
            
            \For{batch $b \in \beta$} {
            
            {$ w \longleftarrow w - \gamma\nabla l(w;b);$ }
            
            }
        
            }
            
        \vspace{3pt}
        
        When ready for an update, pull global model version $ver$ from the server
        
        \vspace{3pt}

        \uIf {$ver - ver_k > a_u$} {
        
        // Client version is too old
        
        Fetch w, ver from the server
        
        \textbf{continue}
        
        \vspace{3pt}
        
        }
        
        \uElseIf {$ver - ver_k < a_l$} {
        
        // Client version is too close to the global
        
        \textbf{continue}
        
        }
        
        \uElse{
        
        \textbf{return} $w$, $ver$ to server

        }

        }
        
        \vspace{5pt}
        
}
\textbf{End Function}
\vspace{5pt}

\caption{Asynchronous Federated Learning: In the system, total $K$ edge vehicles are indexed by $k$; B is the local mini-batch size; E represents the number of local epochs, and $\gamma$ is the learning rate.}
\label{alg-1}
\end{algorithm}

This section describes the algorithm and method used in this article. The diagram of the learning process in a single edge vehicle is shown in Figure \ref{fig:train_edge}. Images are firstly stored in a fixed-sized storage window in order to conduct real-time end-to-end learning based on the continuous image stream. When the storage window reaches its size limit, the most recent picture frames are moved into the training window, while an equivalent number of old frames are dropped. (In our case, the storage window is 100 images wide and the training window is 2,000 wide. These values provide us with the highest model prediction accuracy.) The optical flow information is measured at the same time. Inside the training window, image frames and optical flow frames are fed into a convolutional neural network. The network's performance is compared to the ground truth for that picture frame, which is the human driver's recorded steering wheel angle. Back-propagation is used to adjust the weights of the convolutional neural network in order to enforce the model output as close to the target output as possible.

Following the completion of each training epoch, local models in edge vehicles will be updated to the aggregation server, forming a continuous global awareness among all participating edge vehicles. The following are the steps of the algorithm used in this paper (Algorithm \ref{alg-1}):

\vspace {5pt}
\begin{enumerate}[Step 1:]

\item Edge vehicles compute the model locally; after completing each local training epoch, they retrieve the global model version and compare it to their local version. The decision is based on the frequency bound limits ($a_l$ and $a_u$) and the model version difference $ver$ (global model version) and $ver k$ (local model version of edge vehicle $k$). The upper limit of the model version difference is represented by $a_u$, while the lower limit is represented by $a_l$. There are three conditions:

\end{enumerate}

\begin{itemize}

\vspace{3pt}

\item If the local version is out of date (the client version is too old), the edge vehicle can retrieve the most recent model and conduct local training again.

\vspace{3pt}

\item If the local version is too similar to the latest version (Client is too active), it should stop upgrading and re-train locally.

\vspace{3pt}

\item Clients should then submit modified model results to the aggregation server if the local version is between the upper and lower limits.

\end{itemize}

\vspace{5pt}

\begin{enumerate}[Step 2:]

\item In order to form a global awareness of all local models, the central server performs aggregation based on the ratio determined by the global and local model versions.

\end{enumerate}

\begin{enumerate}[Step 3:]

\item The aggregation server returns the aggregated result to the edge vehicles that request the most recent model.

\end{enumerate}

\vspace{5pt}

Since the algorithm is push-based, the aggregation server only deploys the global model if the edge vehicles request it. When the edge vehicles update their local models, the server aggregates them based on the local model version. The older the model version, the lower the ratio when shaping the global model. Furthermore, although the model update frequency is entirely dependent on local hardware settings, there are two bound limits in place to ensure that the update frequency of local clients is within a reasonable range $[a_l, a_u]$. (In our case, based on the number of the participated vehicles, the lower frequency bound $a_l$ we set equals to 2 while the upper bound $a_u$ is 6.)

\begin{figure*}[!t]%
\centering
\subfigure[][Vehicle 1]{%
\label{fig:ex3-a}%
\includegraphics[height=2.2in]{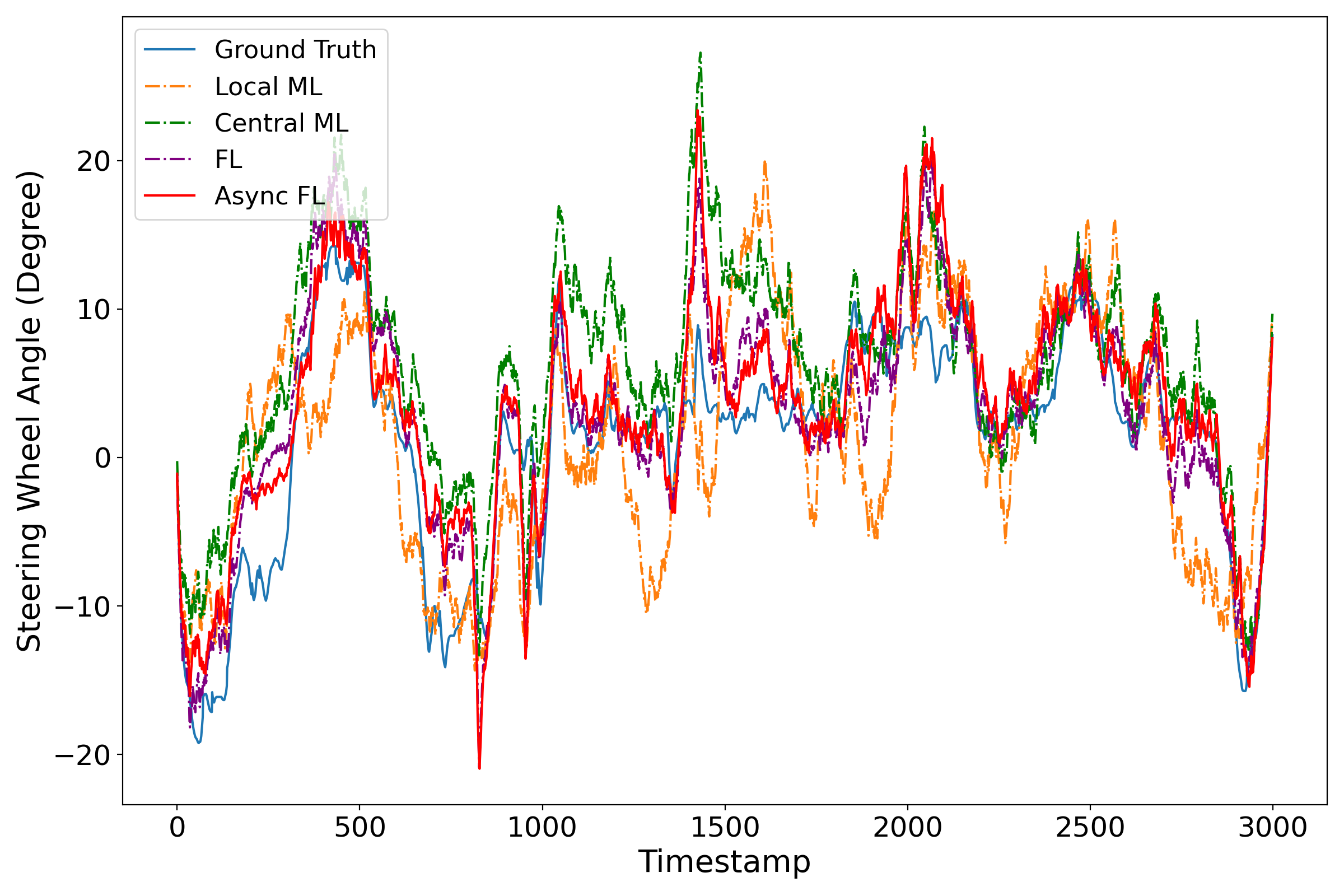}}%
\hspace{16pt}%
\subfigure[][Vehicle 2]{%
\label{fig:ex3-b}%
\includegraphics[height=2.2in]{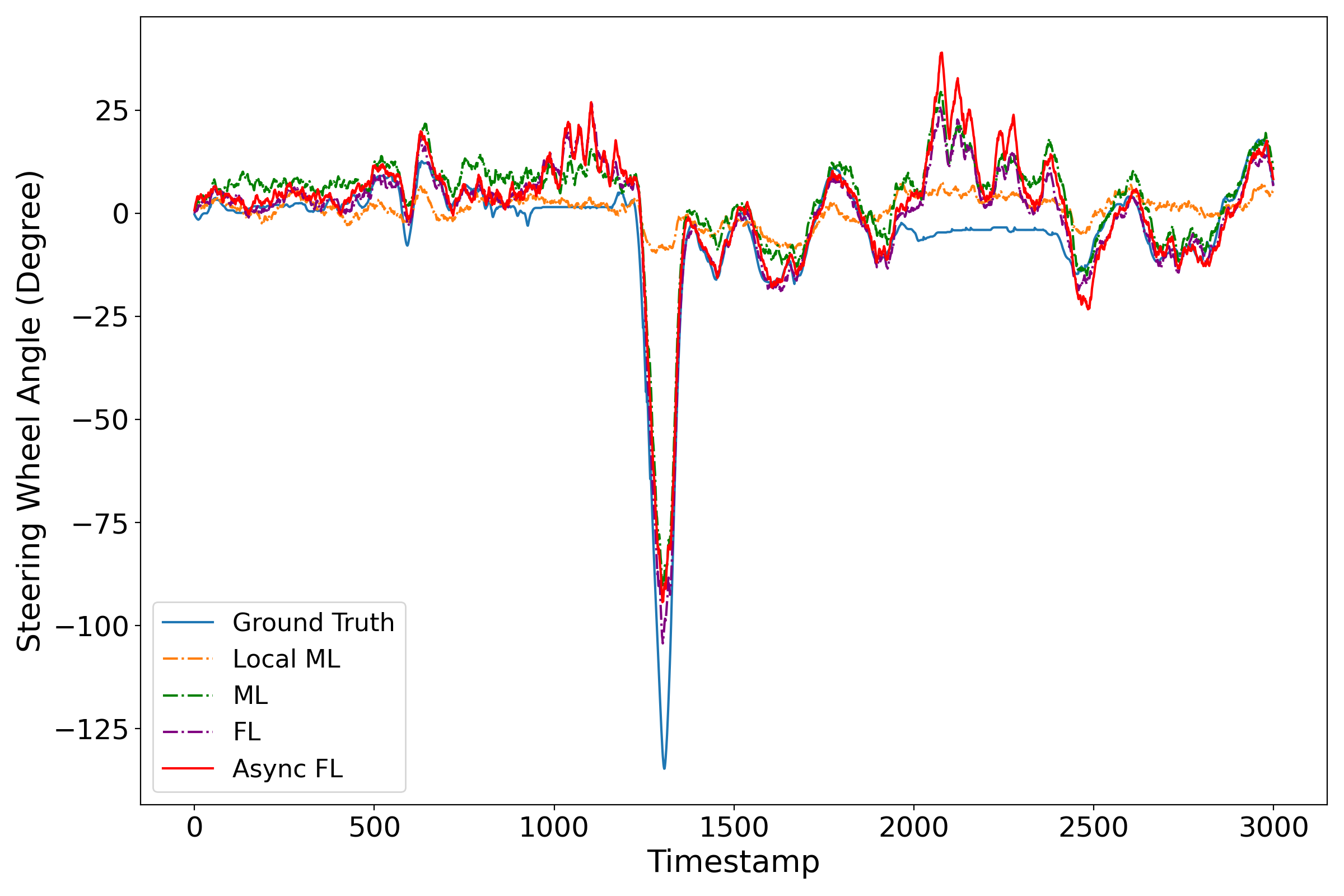}} \\
\subfigure[][Vehicle 3]{%
\label{fig:ex3-c}%
\includegraphics[height=2.2in]{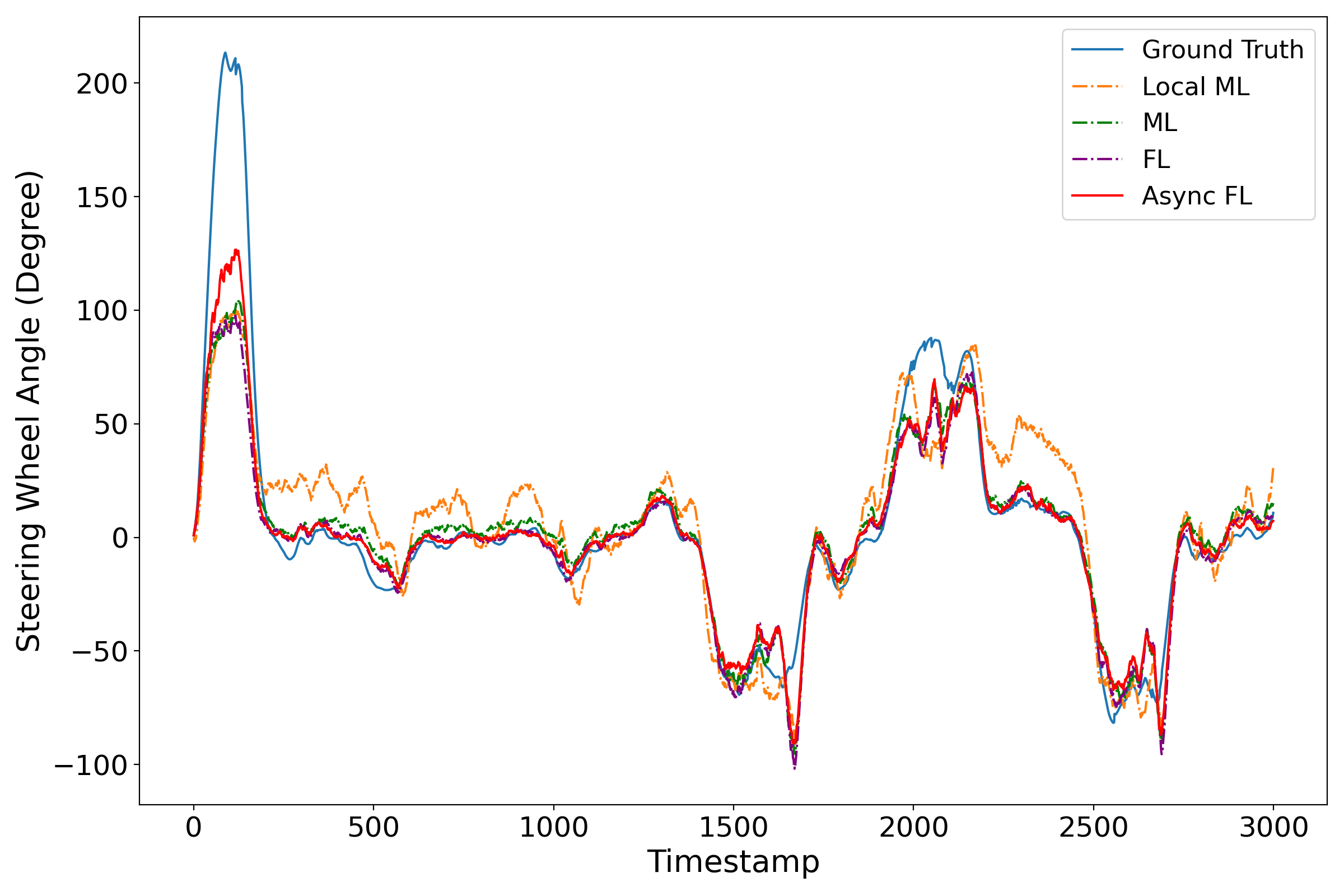}}%
\hspace{16pt}%
\subfigure[][Vehicle 4]{%
\label{fig:ex3-d}%
\includegraphics[height=2.2in]{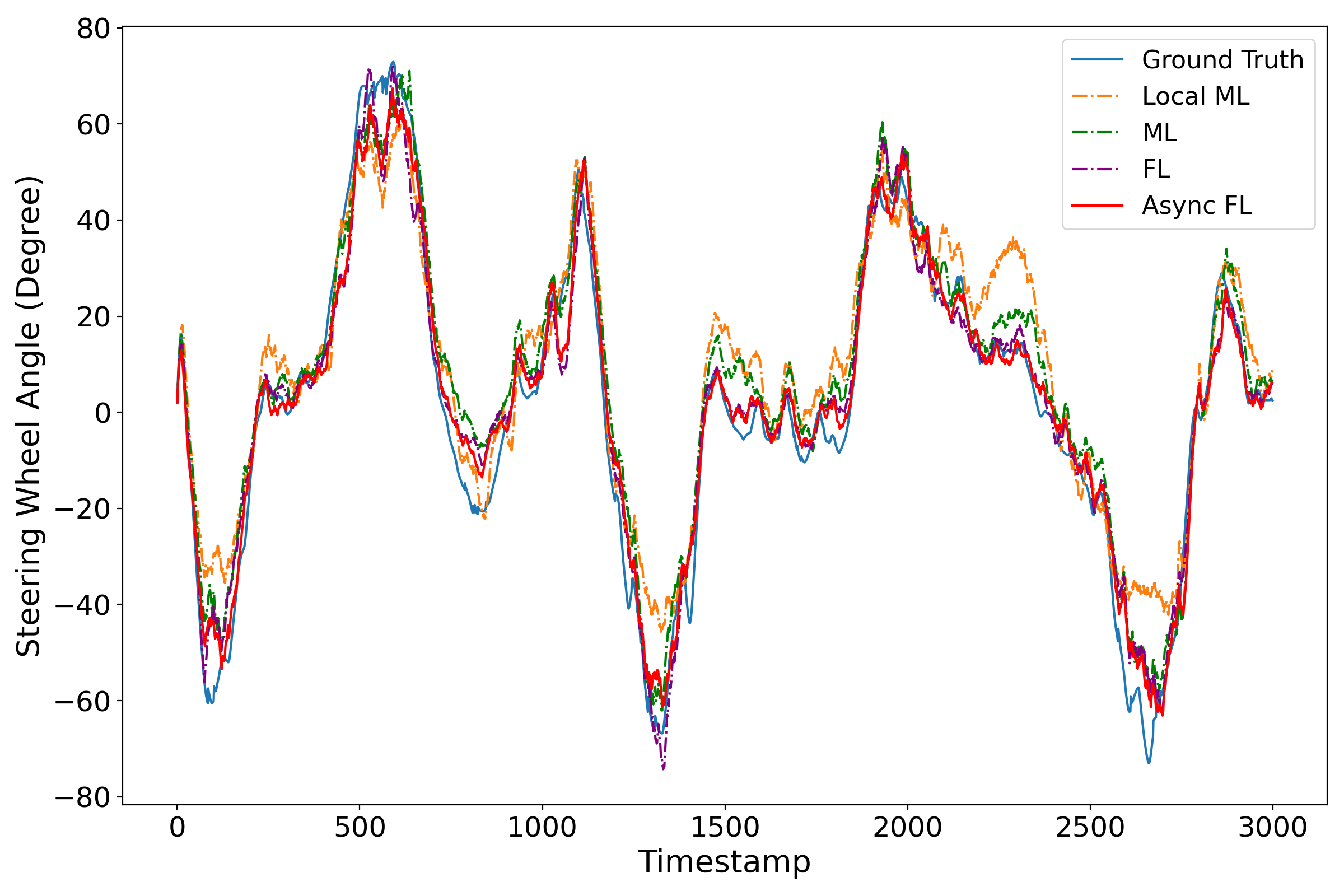}}%
\caption[]{The comparison of angle prediction performance on four local vehicle test set with Federated Learning and three baseline models.}%
\label{fig:result}%
\end{figure*}

\begin{figure}[h]%
\centering
\subfigure[][Vehicle 1]{%
\label{fig:ex3-a}%
\includegraphics[height=1.1in]{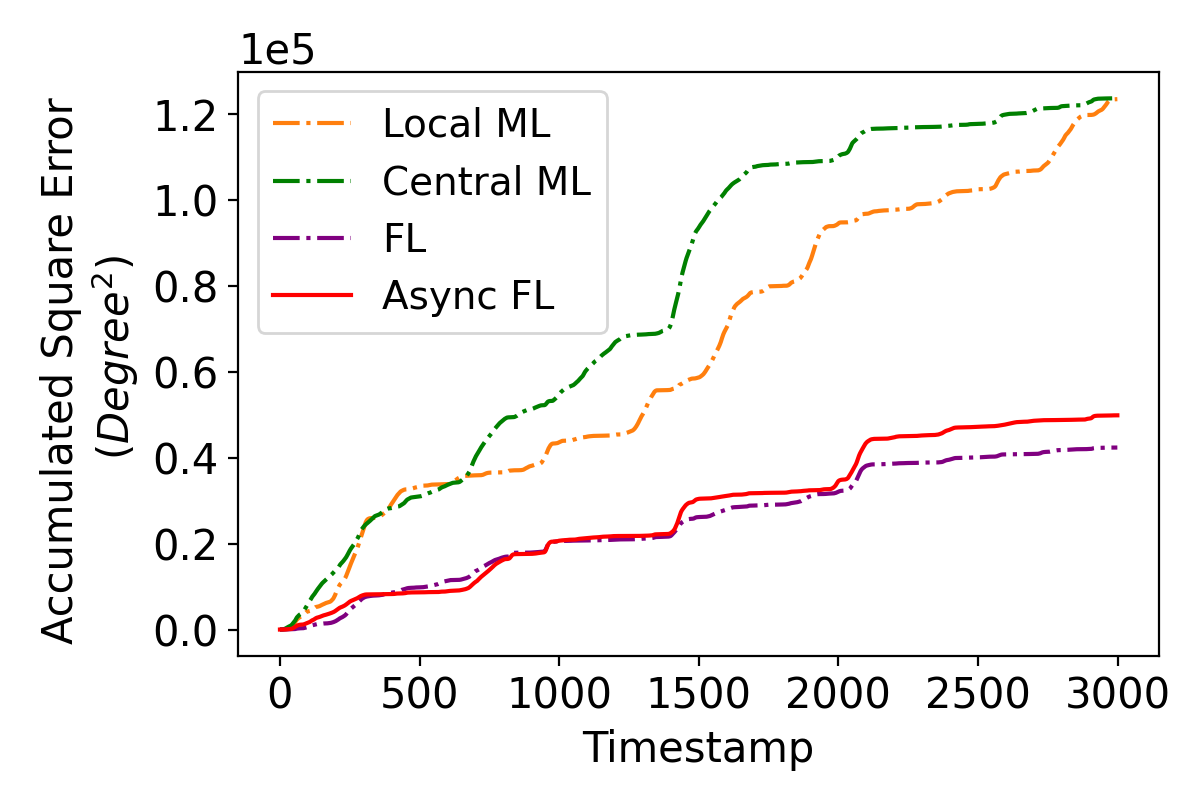}}
\subfigure[][Vehicle 2]{%
\label{fig:ex3-b}%
\includegraphics[height=1.1in]{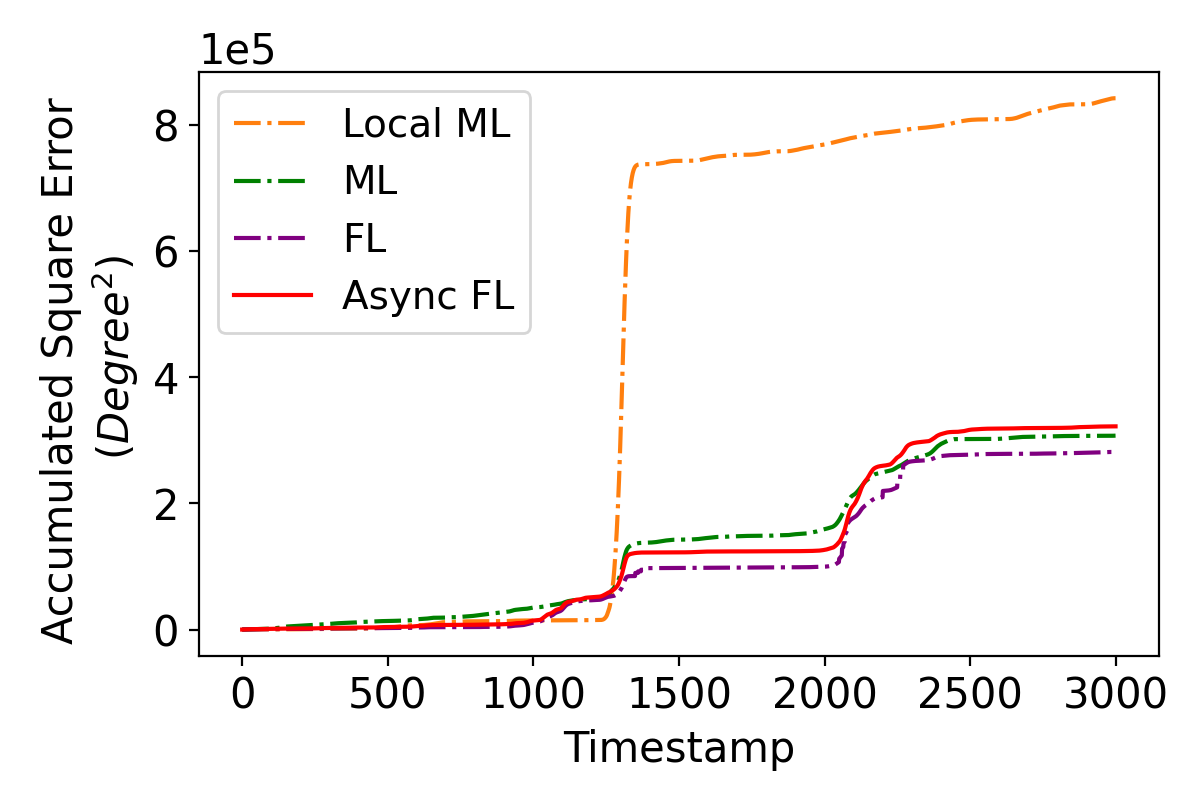}} \\
\subfigure[][Vehicle 3]{%
\label{fig:ex3-c}%
\includegraphics[height=1.1in]{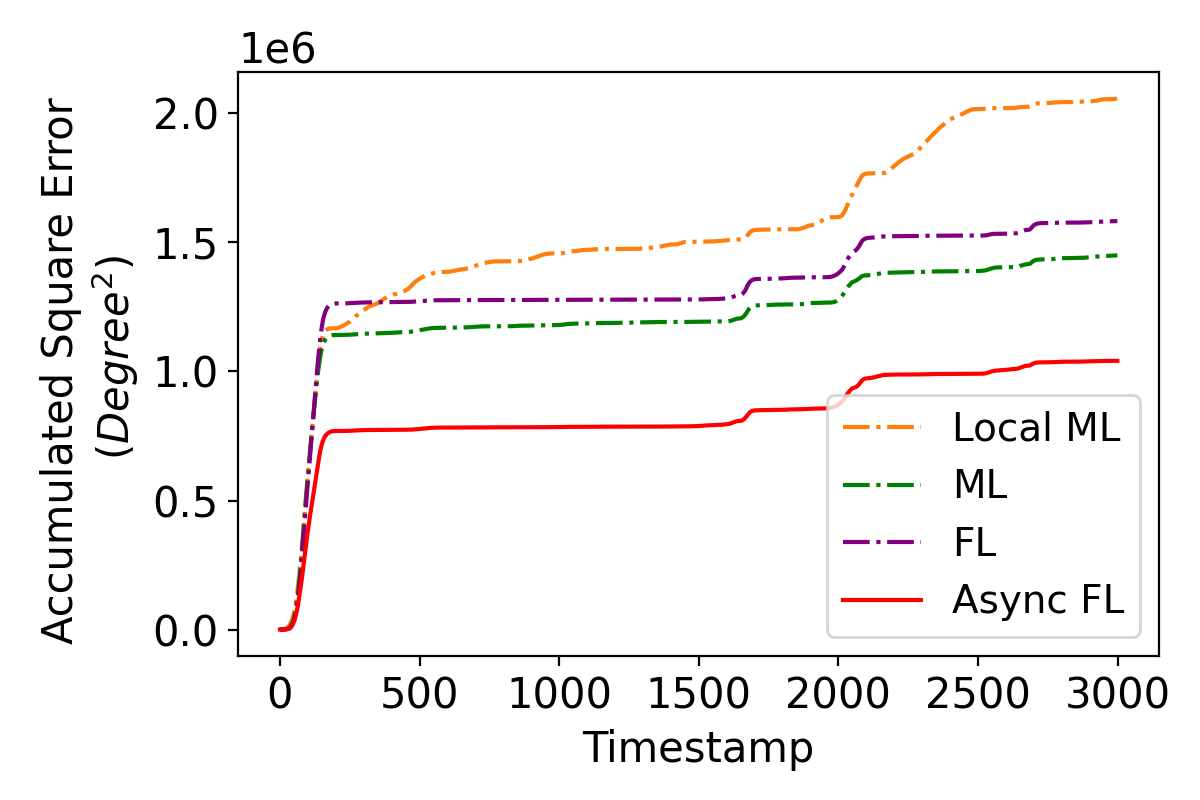}} 
\subfigure[][Vehicle 4]{%
\label{fig:ex3-d}%
\includegraphics[height=1.1in]{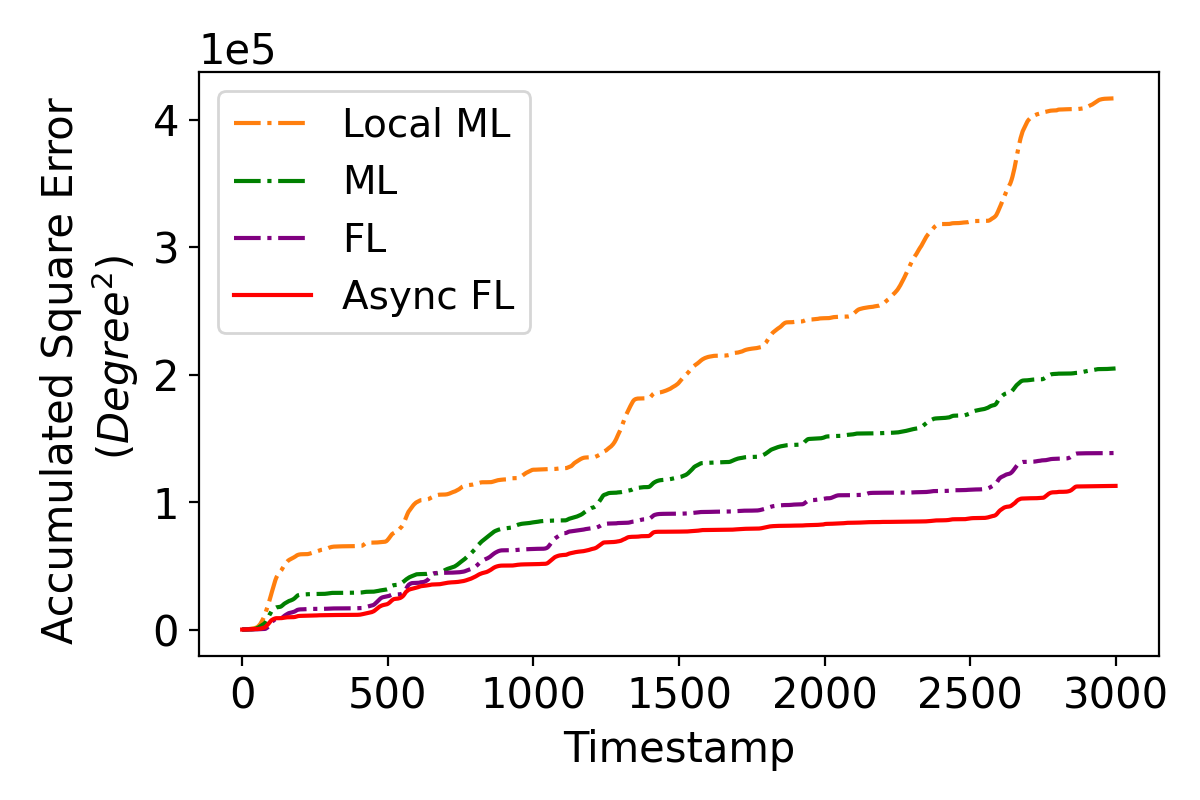}}%
\caption[]{Accumulated error on test dataset in 4 edge vehicles with asynchronous Federated Learning and other baseline models.}%
\label{fig:accu-result}%
\end{figure}

\section{Results}
\label{sec:eval}

We present the experiment results of the real-time end-to-end Federated Learning approach to steering wheel angle prediction in this section. The device output is evaluated based on three factors, as defined in Section \ref{sec: method}. (The metrics are described in \ref{sec:metric}.) - (1) Angle prediction performance (2) Model Training Time (3) Bandwidth cost. The results are compared with other three baseline models which are trained by - 1) Traditional Centralized Learning (ML) 2) Local training without model sharing (Local ML) 3) Synchronous Federated Learning (FL)

Figure \ref{fig:result} compares the angle prediction output of the model trained by asynchronous Federated Learning (Async FL) to the other baseline models. The results show that the Federated Learning models (synchronous and asynchronous) may achieve the same or even better prediction accuracy than the traditional centralized trained model. The Federated Learning model reacts faster than other learning approaches, particularly at the timestamps that require rapid changes in steering wheel angle. Furthermore, when compared to independently trained models, Federated Learning approaches can provide a much better prediction that is much closer to the ground truth.

To provide a clear view of model output with different approaches, we accumulated the square error between expected angle and ground truth (calculated by $(\theta_t-\hat{\theta_t})^2$) and demonstrate it in Figure \ref{fig:accu-result}. The results provide the same information as Figure \ref{fig:result}. We find that asynchronous Federated Learning outperforms centralized learning and local machine learning. In addition, as compared to synchronous Federated Learning, our method achieves higher prediction accuracy in edge vehicles 3 and 4. Table \ref{tab:result} displays detailed numerical results, including the regression error (RMSE) on each test dataset in each vehicle and the overall average accuracy among the test datasets of all participating edge vehicles. 

\begin{figure*}[!t]%
\centering
\subfigure[][Vehicle 1]{%
\label{fig:ex3-a}%
\includegraphics[height=1.6in]{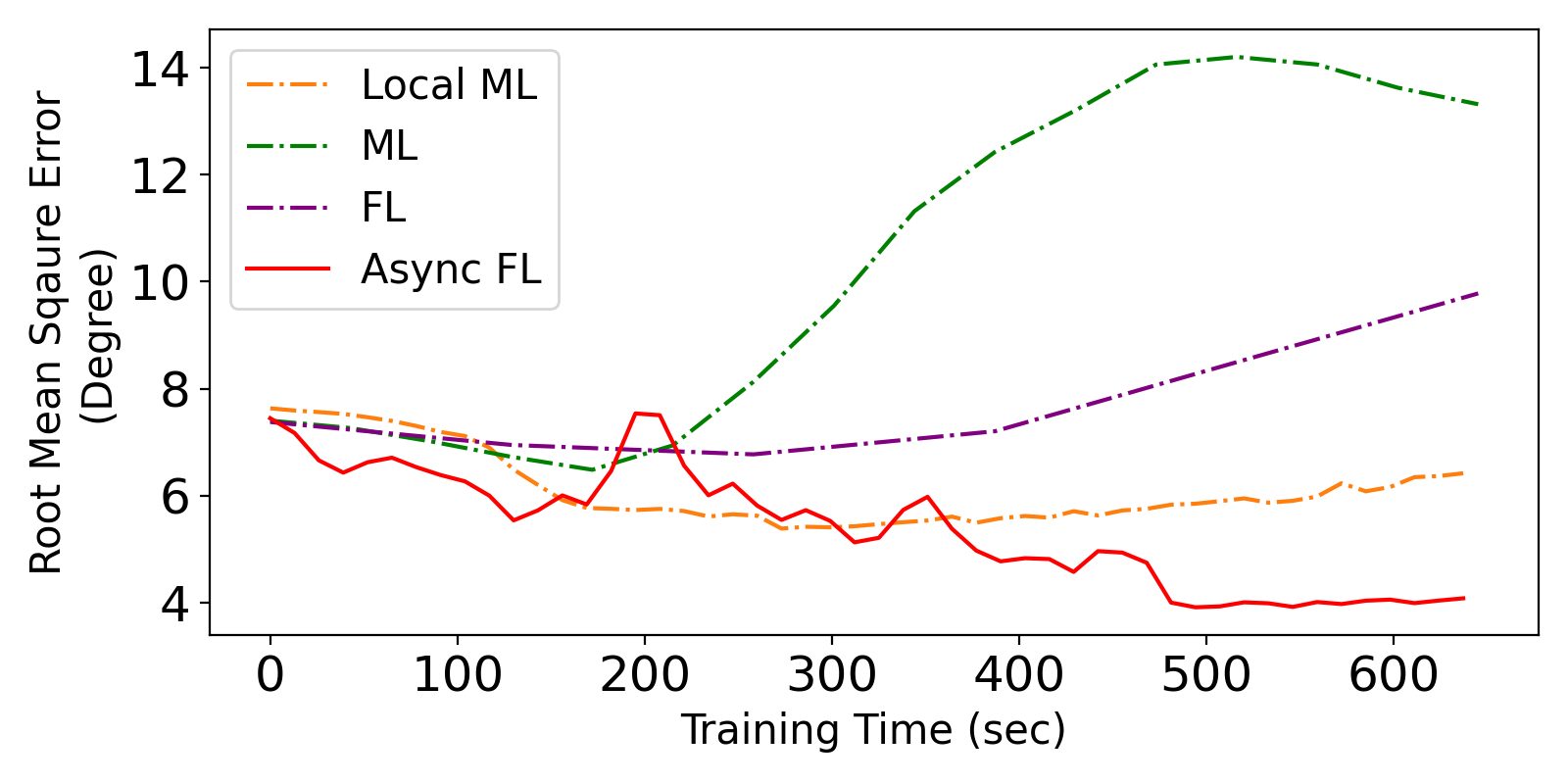}}%
\hspace{8pt}%
\subfigure[][Vehicle 2]{%
\label{fig:ex3-b}%
\includegraphics[height=1.6in]{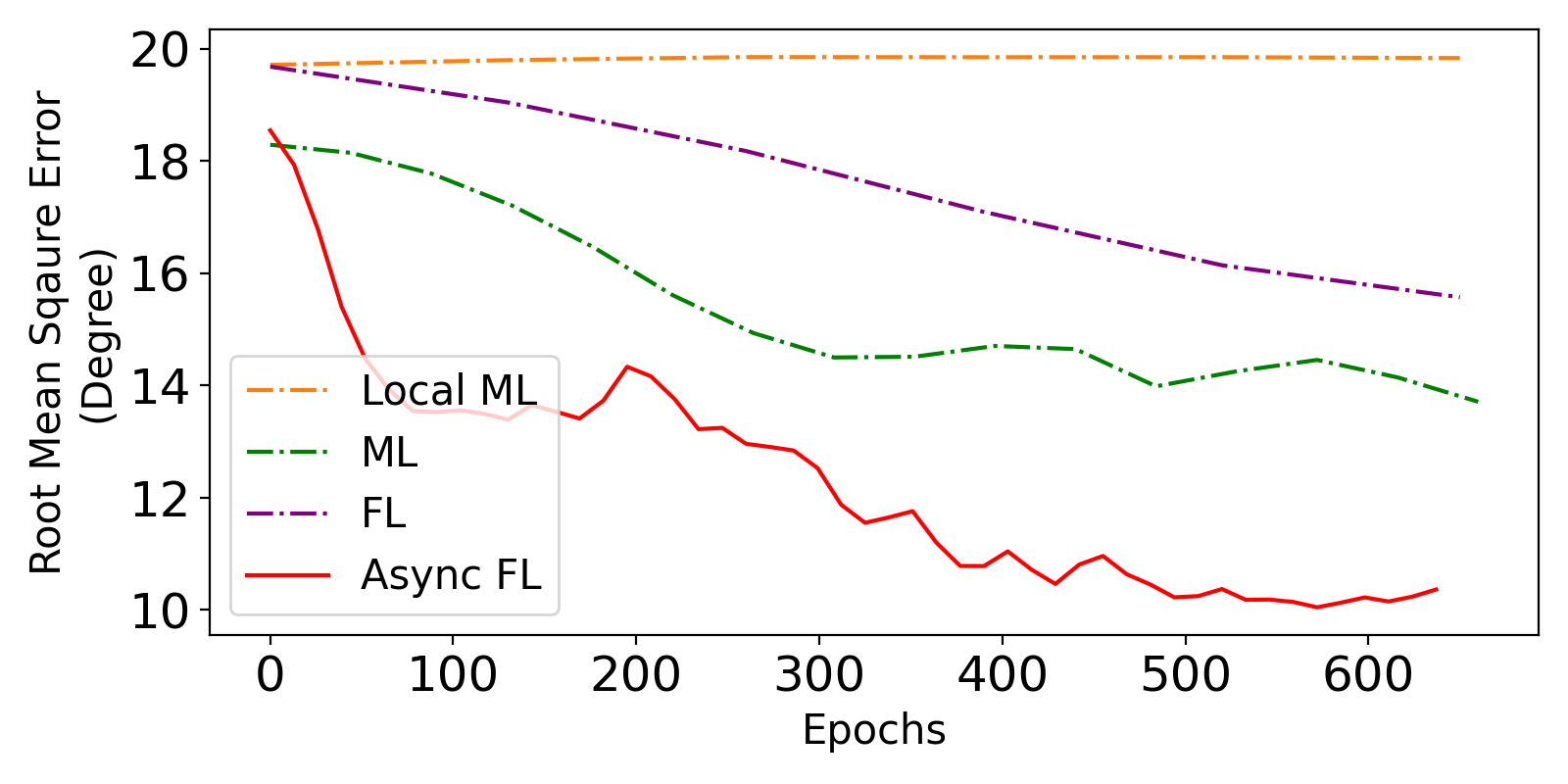}} \\
\subfigure[][Vehicle 3]{%
\label{fig:ex3-c}%
\includegraphics[height=1.6in]{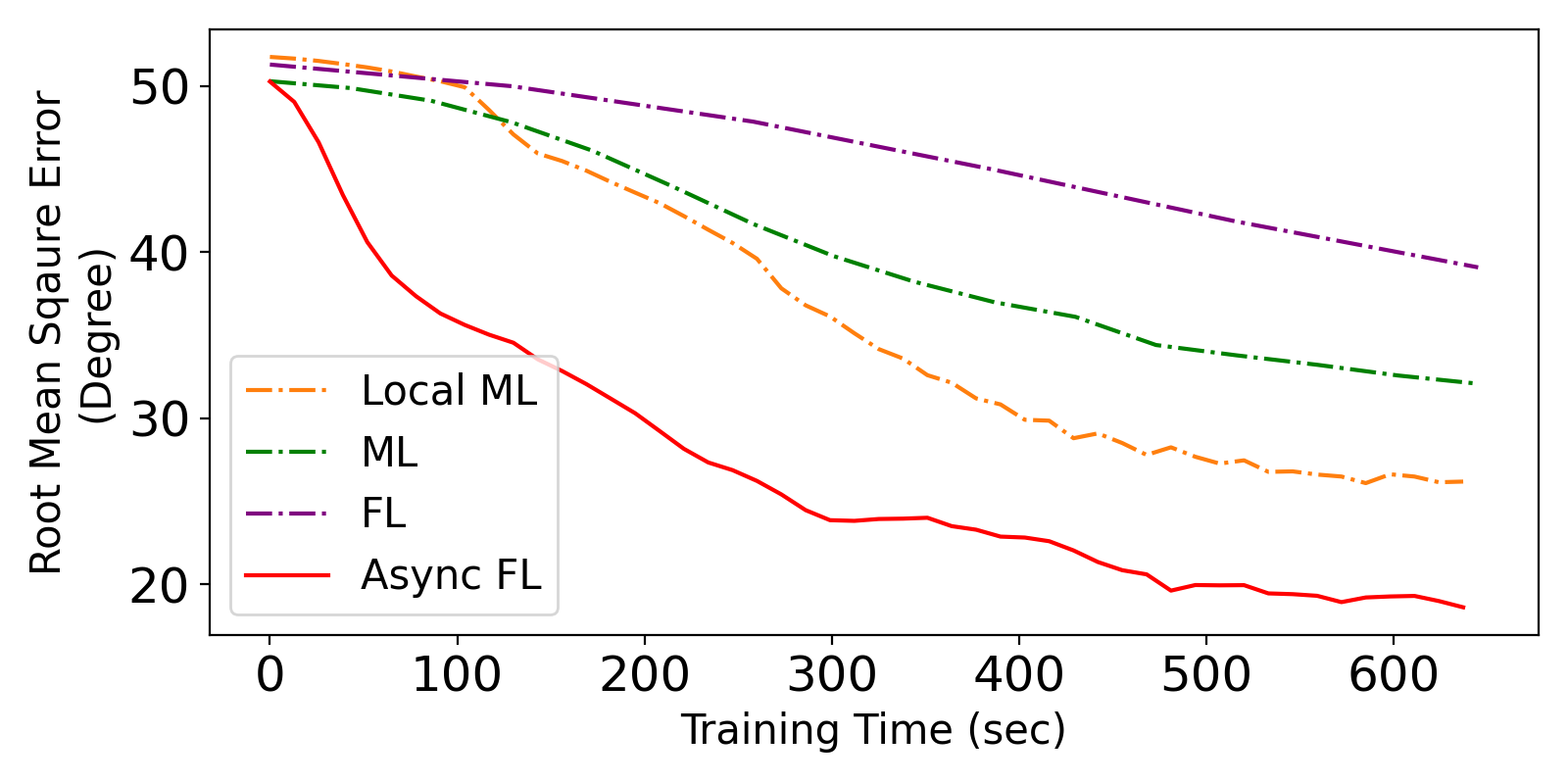}}%
\hspace{8pt}%
\subfigure[][Vehicle 4]{%
\label{fig:ex3-d}%
\includegraphics[height=1.6in]{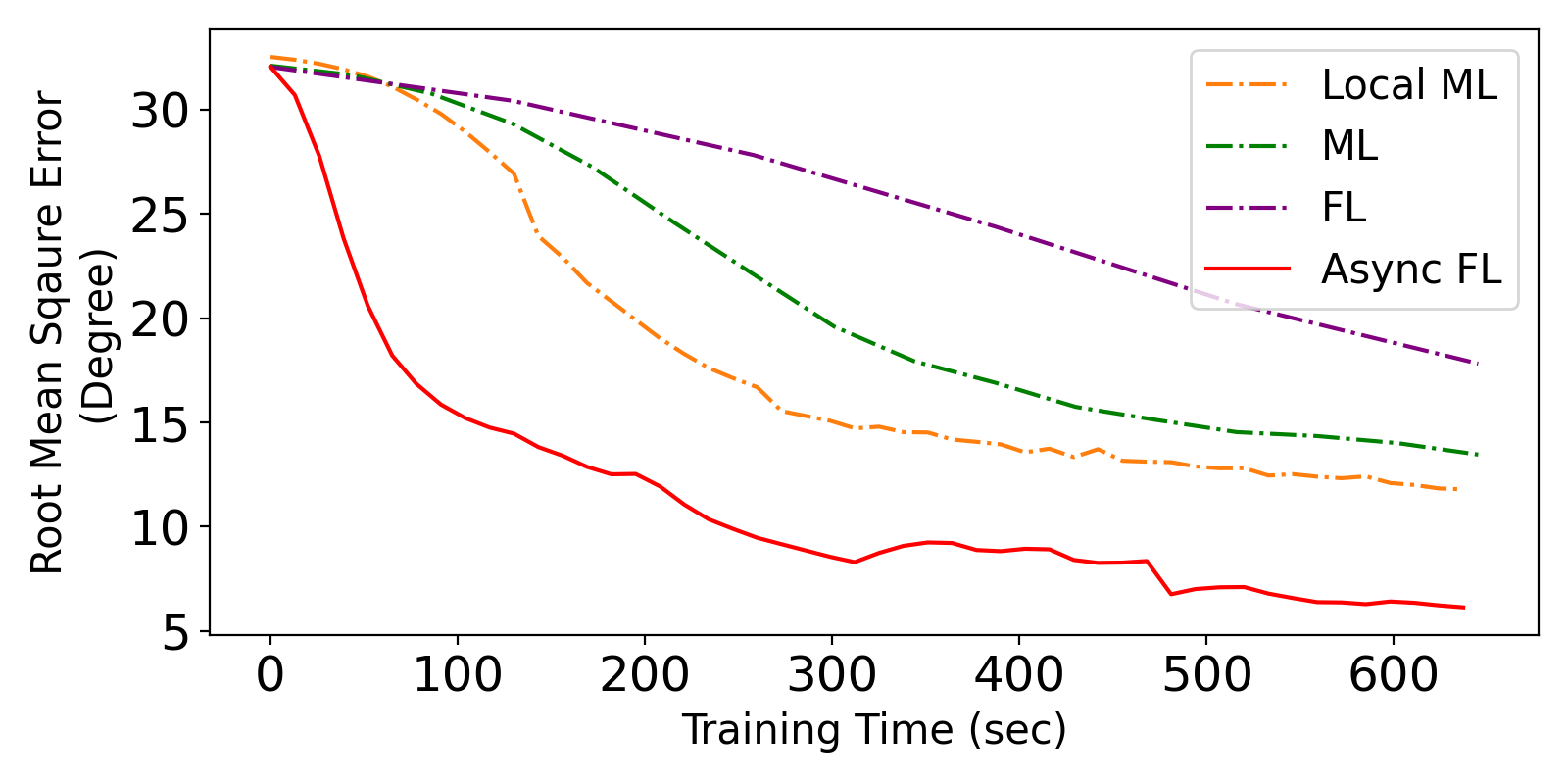}}%
\caption[]{The comparison between angle prediction performance and the model training time on four local vehicle test set with asynchronous Federated Learning and three baseline models.}%
\label{fig:con-result}%
\end{figure*}

\begin{table}[h]
 \caption{Steering wheel angle regression error (RMSE) on test set of each edge vehicle (4 vehicles in total)}
\label{tab:result}
\centering
 \begin{tabular}{c c c c c c} 

\toprule
 & Vehicle 1 & Vehicle 2 & Vehicle 3 & Vehicle 4 & Overall \\
 \midrule
 Async FL & 4.077 & 10.358 & \textbf{18.629} & \textbf{6.129} & \textbf{11.275}\\
 FL & 3.758 & 9.933 & 22.967 & 6.795 & 12.754\\ 
 ML & 6.422 & 10.118 & 21.985 & 8.264 & 13.183\\
 Local ML & 6.416 & 16.749 & 26.196 & 11.788 & 16.954\\
\bottomrule
 \end{tabular}

\end{table}

The findings show that asynchronous Federated Learning outperforms other baseline models in vehicles 3 and 4. In vehicle 1 and 2, models trained by asynchronous Federated Learning only perform about $0.2$ and $0.4$ worse than the synchronous Federated learning method. Based on our findings, we may conclude that the asynchronous Federated Learning model can provide better prediction performance than the local independently trained model, and its behaviour can achieve the same or even higher accuracy level when compared to centralized learning and the synchronous Federated Learning model.

Furthermore, Figure \ref{fig:con-result} illustrates the shift in regression error with model training time in order to evaluate model training efficiency. The results show that the asynchronous Federated Learning method outperforms all of the baseline approaches in terms of model training efficiency. With the same training period, our approach can achieve better prediction efficiency (with approximately 50\% less regression error) and converge approximately 70\% faster than other baseline models.

\begin{table}[]
\centering
 \caption{Total Training Time and Bandwidth cost with different model training methods (4 Vehicles in total)}
\label{tab:t_result}
 \begin{tabular}{c c c c c} 

\toprule
 & Async FL & FL & ML & Local ML \\
\midrule
Training Time (sec) & \textbf{669.2} & 5,982.8 & 2143.7 & 5,903.4\\ 
Bytes Transferred (GB) & \textbf{0.78} & 0.78 & 2.02 & - \\
\bottomrule
\end{tabular}

\end{table}

The comparison of total training time and bytes transferred between Federated Learning and three baseline models is shown in table \ref{tab:t_result}. For all the models, the total number of training epochs is 50. With async FL, FL, and Local ML learning approaches, model training is accelerated by Nvidia Tesla V100 GPU in edge vehicle 1, while model training is accelerated by Nvidia Tesla T4 GPU in edge vehicle 3, 4. The ML method completes training on a single server with Nvidia Tesla T4 GPU acceleration. As compared to the traditional centralized learning approach, the bandwidth cost of both Federated Learning methods is reduced by approximately 60\%. The results for model training time indicate that asynchronous Federated Learning needs significantly less training time than other baseline methods. However, since there is no GPU available for synchronous Federated Learning and local learning, edge vehicle 2 becomes the burden of the entire system. Other vehicles must wait for vehicle 2 to complete its local training round before performing model aggregation and further training tasks, which is inflexible and time-consuming. The performance of these two methods is even lower than that of the centralized learning system with GPU acceleration. In summary, as compared to the traditional centralized learning process, asynchronous Federated Learning reduces training time by approximately 70\% and saves approximately 60\% bandwidth. Since our method consumes real-time streaming data, there is no need to store and train on a large dataset in a single edge unit, making it cost-effective and relevant to real-world systems.

\section{Conclusion and Future Work}
\label{sec:concl}

In this paper, we present a novel approach to real-time end-to-end Federated Learning using a version-based asynchronous aggregation protocol. We validate our approach using a critical use case, steering wheel angle prediction in self-driving cars. Our findings show the model's strength and advantages when trained using our proposed method. In our case, the model achieves the same or even better prediction accuracy than widely used centralized learning methods and other Federated Learning algorithms while reducing training time by 70\% and bandwidth cost by 60\%. Note that the decrease would be more visible if the number of participating devices is expanded more, which proves to be cost-effective and relevant to real-world systems.

In the future, we plan to further analyze our algorithm with different combinations of hyper-parameters, such as the aggregation frequency bound $a_l$ and $a_u$. As the parameter settings become more important with the number of participating learning vehicles increases, we would like to add more federated edge users in order to test device output that may differ with these bounds. In addition, we will test our approach in additional use cases and investigate more sophisticated neural networks combined with our approach. In addition, we plan to develop more appropriate aggregation algorithms and protocols in order to increase model training performance on resource-constrained edge devices in real-world embedded systems.

\section*{Acknowledgement}

This work was funded by the Chalmers AI Research Center. The computations were enabled by resources provided by the Swedish National Infrastructure for Computing (SNIC) at Chalmers Centre for Computational Science and Engineering (C3SE), which is partially funded by the Swedish Research Council through grant agreement no. 2018-05973. The authors would also like to express their gratitude for all the support and suggestions provided by colleagues from Volvo Cars.

\bibliographystyle{./bibliography/IEEEtran}
\bibliography{cite}

\end{document}